\documentclass[conference]{IEEEtran}
\IEEEoverridecommandlockouts
\usepackage{cite}
\usepackage{amsmath,amssymb,amsfonts}
\usepackage{algorithmic}
\usepackage{graphicx}
\usepackage{textcomp}
\usepackage{xcolor}
\usepackage{multicol}
\usepackage{subcaption}
\usepackage{booktabs}
\usepackage{comment}

\def\BibTeX{{\rm B\kern-.05em{\sc i\kern-.025em b}\kern-.08em
    T\kern-.1667em\lower.7ex\hbox{E}\kern-.125emX}}
\begin{document}

\title{Deep Heterogeneous Autoencoders \\
for Collaborative Filtering\\
}

\author{
    \IEEEauthorblockN{
    	Tianyu Li\IEEEauthorrefmark{1}\thanks{Correspondence to: tianyu.li@rakuten.com}, 		 Yukun Ma\IEEEauthorrefmark{2}, Jiu Xu\IEEEauthorrefmark{1}, 
        Bj\"orn Stenger\IEEEauthorrefmark{1}, Chen Liu\IEEEauthorrefmark{1}, 
        Yu Hirate\IEEEauthorrefmark{1}}
    \IEEEauthorblockA{\IEEEauthorrefmark{1}Rakuten Institute of Technology
    }
    \IEEEauthorblockA{\IEEEauthorrefmark{2}Nanyang Technological University
    }
}

\maketitle

\begin{abstract}
This paper leverages heterogeneous auxiliary information to address the data sparsity problem of recommender systems.
We propose a model that learns a shared feature space from heterogeneous data, such as item descriptions, product tags and online purchase history, to obtain better predictions. Our model consists of autoencoders, not only for numerical and categorical data, but also for sequential data, which enables capturing user tastes, item characteristics and the recent dynamics of user preference. 
We learn the autoencoder architecture for each data source independently in order to better model their statistical properties.
Our evaluation on two {\em MovieLens} datasets and an e-commerce dataset shows that mean average precision and recall improve over state-of-the-art methods.
\end{abstract}

\begin{IEEEkeywords}
Deep Autoencoder, Heterogeneous Data, Shared Representation, Sequential Data Modeling, Collaborative Filtering
\end{IEEEkeywords}

\section{Introduction}
\begin{figure}
  \centering
  \includegraphics[width=\columnwidth]{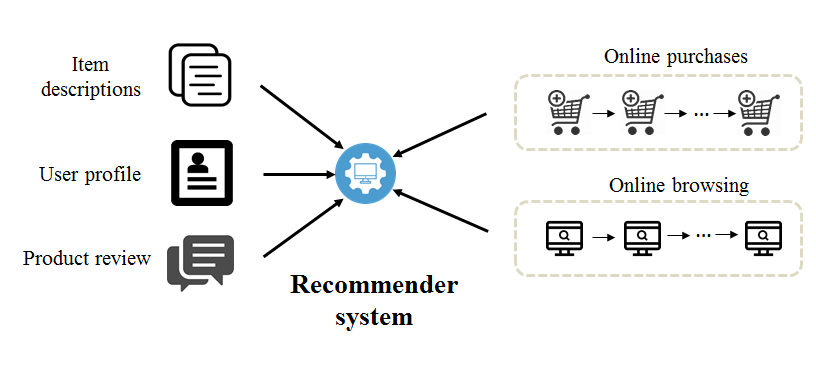}
  \caption{{\bf Auxiliary information usage in recommender systems.} \it Item descriptions and user profiles are typically being used for feature extraction to alleviate the data sparsity problem. 
  Our proposal also leverages sequential data, such as purchase histories, to reflect user preferences.}
  \label{fig:auxiliary_usage}
\end{figure}

Although Collaborative Filtering (CF) techniques achieve good performance in many recommender systems~\cite{Hu2008}, their performance degrades significantly when historical data is sparse.
In order to alleviate this problem, features from auxiliary data sources that reflect user preference have been extracted~\cite{Oord2013,Porteous2010}, as shown in Fig. \ref{fig:auxiliary_usage}. How to represent data from different sources is still a research problem, and it has been shown that the representation itself substantially impacts performance~\cite{Loyola2017,Goodfellow2016}. Recently, representation learning that automatically discovers hidden factors from raw data has become a popular approach to remedy the data sparsity issue of recommender systems~\cite{wangweiran2015,Zheng2017}.

Many online shopping platforms gather not only user profiles and item descriptions, but various other types of data, such as product reviews, tags and images. Recent research has added textual and visual information to recommender systems~\cite{Fuzheng2016,Oramas2017}. 
However, in many cases sequential data, such as user purchase and browsing history, which carries information about trends in user tastes, have largely been neglected in CF-based recommender systems. 

In this paper we propose Deep Heterogeneous Autoencoders (DHA) for Collaborative Filtering to combine information from multiple domains. We use Stacked Denoising Autoencoders (SDAE) to extract latent features from non-sequential data, and Recurrent Neural Network Encoder-Decoders (RNNED) to extract features from sequential data.
The model is able to capture both  user preferences and potential shifts of interest over time.
Each data source is modeled using an independent encoder-decoder mechanism.  
Different encoders can have different number of hidden layers and an arbitrary number of hidden units in order to deal with the intrinsic difference of data sources. For instance, user demographic data and item content are typically categorical, while user comments or item tags are textual. After pre-processing, such as one hot encoding, bag-of-words and word2vec computation, representation vectors are on a different level of abstraction. Owing to its flexible structure, our model is able to learn suitable latent feature vectors for each component.  
These local representations from each data source are joined to form a shared feature space, which couples the joint learning of the representation from heterogeneous data and the collaborative filtering of user-item relationships. 

The contributions of this paper are summarized as follows:
\begin{enumerate}
\item  A method for modeling both static and sequential data in a consistent way for recommender systems in order to capture the trend in user tastes, and 
\item Adaptation of the autoencoder architecture to accurately model each  data source by considering their distinct abstraction levels. 
\end{enumerate}
We show improvements in terms of mean average precision and recall on three different datasets.

\section{Related work}

\subsection{Incorporating side information into recommender systems}

In order to improve recommendation performance, research has been focusing on using side information, such as user profiles and reviews~\cite{Fuzheng2016, Porteous2010}. In particular, deep learning models have been widely studied~\cite{He2017,Wu2017}.
AutoRec first proposed the use of autoencoders for recommender systems~\cite{Sedhain2015}. In more recent work, representations are learned via  stacked autoencoders (SAE), and fed into conventional CF models, either loosely or tightly coupled~\cite{szhang2017, hwang2015}. Deep models that integrate autoencoders into collaborative filtering have shown state-of-the-art performance. 


\subsection{Recurrent Neural Network Encoder-Decoder}

Recurrent neural networks~(RNNs) process sequential data one element at each step to capture temporal dynamics. The encoder-decoder mechanism was initially applied to RNN for machine translation~\cite{Cho2014}. Recently, RNN encoder-decoders~(RNNED) have been used to learn features from a series of actions and have successfully been applied in other areas. It was shown that Long Short-Term Memory (LSTM) networks have the ability to learn on data with long range temporal dependencies, and we adopt LSTMs for modeling sequential data.

\begin{figure}
  \centering
  \includegraphics[width=0.45\textwidth]{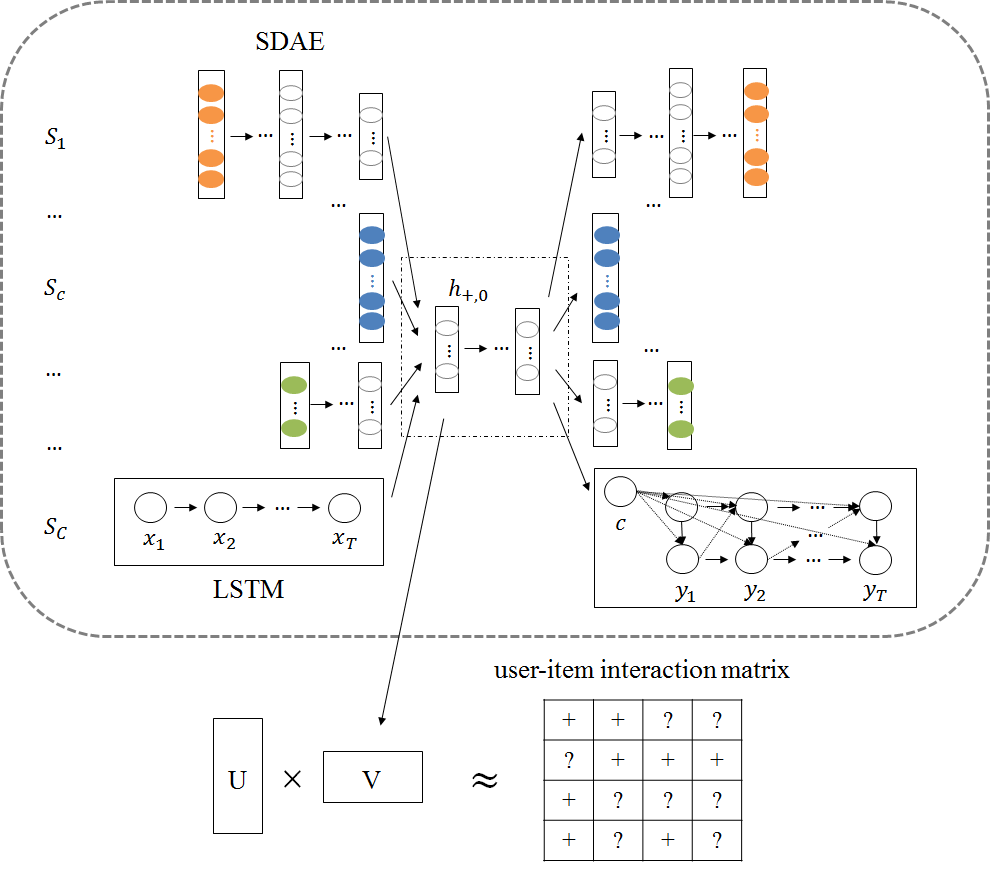}
  \caption{{\bf Deep Heterogeneous Autoencoders and the integration with collaborative filtering.} \it The proposed model extracts a shared feature space from multiple sources of auxiliary information. It models non-sequential and sequential data to capture user preferences, item properties as well as temporal dynamics. It adopts independent encoder-decoder architectures for different data sources in order to better model their statistical properties. The  product of $U \in \mathbb{R}^{m \times d}$ and $V \in \mathbb{R}^{n \times d}$ approximates the user-item interaction matrix.}
  \label{fig:DHA}
\end{figure}


\section{Deep Heterogeneous Autoencoders for Collaborative Filtering}

\subsection{Overview} 

We propose a model that learns a joint representation from heterogeneous auxiliary information to mitigate the data sparsity problem of recommender systems. SDAEs are applied to numerical and categorical data for modeling the static tastes of users for items. We use RNNEDs to extract features from sequential data to reveal interest shifts over time. 

The model adopts an independent autoencoder architecture for each data source since the inputs are generally on a different level of abstraction, see Fig.~\ref{fig:DHA} for an overview. 
In order to discover the distinct statistical properties of every data source, our model takes the existing disparity of input abstraction levels into consideration, and applies autoencoders to each source independently by allowing distinct hidden layer numbers and arbitrary hidden units at every layer.



\subsection{Deep Heterogeneous Autoencoders}


We define each source of auxiliary data as a component indexed by $c \in \{1,...,C\}$. $S_c$ denotes the input of component $c$. We pre-process non-sequential data like textual item descriptions by generating fixed-length embedding vectors. For sequential data, an embedding vector is learned for every time step after tokenization. We seperately describe the encoding-decoding outputs of the above two types of embedding vectors. 


As shown in Fig. \ref{fig:DHA}, SDAE is applied to fixed-length embedding vectors.
Each component encoder takes the input $S_c$, generates a corrupted version of it, $\hat{S_c}$, and the first layer maps it to a hidden representation $h_c$, which captures the main factors of variation in the input data distribution\cite{vincent2008,vincent2010}.
More importantly, the number of component hidden layers in our model can differ from each other. The architecture is unique for each data source, where the number of layers of component $c$ is denoted as $L_c$. The representation at every layer is $S_{c,l}$. For the encoder of each component, given $l_c \in \{1, ..., L_c/2 \}$ and $c \in C$, the hidden representation $h_{c,l}$ is derived as:
\begin{equation}
h_{c, l} = f \left(W_{c,l}h_{c,l-1} + b_{c,l} \right).
\end{equation}

The decoder reconstructs the data at layer $L$ as follows:
\begin{equation}
\bar{S}_c = g\left( W'_c h_{c,L} + b'_c \right).
\end{equation}



The proposed model leverages sequential data by using two LSTMs for encoding and decoding one sequential data source. Specifically, the encoder reads a sequence with $T$ time steps. At the last time step, the hidden state $h_T$ is mapped to a context vector $c$, as a summary of the whole input sequence\cite{Cho2014}. The decoder generates the output sequence by predicting the next action $y_t$ given $h_t$. Both $y_t$ and $h_t$ are also conditioned on $y_{t-1}$ and the context vector $c$. 


To combine them, as shown in Fig. \ref{fig:DHA}, the first part of our model encodes all components to generate hidden representations $S_{c, L_c/2}$ of non-sequential data and $h_T$ of sequential data across all sources. These are merged to generate a joint latent representation, denoted as $h_{+,0}$.
Analogous to the hidden layers of each component, the fusion model can have multiple hidden layers, the total number denoted as $L_+$. The representation of the first fusion hidden layer is 
\begin{equation}
h_{+,0} = f \left(\sum_{c\in C} W_{c, +}h_{c,L_c/2} + b_{+,0} \right).
\end{equation}

The first hidden layer $h_{+,0}$ of the fused model is fed into the collaborative filtering model. After joint training, $h_{+,0}$ is the latent vector to generate recommendation results.

\subsection{DHA-based Collaborative Filtering}




All data is fed into two DHAs for users and items, respectively. Fig.~\ref{fig:DHA} shows the process for items, and it is analogous for user data. 
Let $R \in \mathbb{R}^{m \times n}$ denote the rating matrix of users to items, $S_c^{\left(u \right)}$ being the component $c$ input for users and $S_c^{\left(v \right)}$ that for items. Then, $h_{+,0}^{\left(u \right)}$ and $h_{+,0}^{\left(v \right)}$ are the latent factors. The loss function of the proposed DHA based collaborative filtering is defined as: 
\begingroup\makeatletter\def\f@size{8}\check@mathfonts
\begin{flalign}
\begin{split}
\label{dha_cost_func}
L = &\sum_{i,j} c_{i,j} \left(r_{i,j} - u_i v_j \right) ^ 2 + \lambda_f (\sum_i||u_i||^2 +  \sum_j ||v_j||^2) \\
      &+ \lambda_{m} \sum_{c\in C_u} \emph{loss}(S_c^{\left(u\right)}, \bar{S}_c^{\left(u\right)}) + \lambda_{n} \sum_{c\in C_i} \emph{loss}(S_c^{\left(v\right)}, \bar{S}_c^{\left(v\right)}) \\
      &+ \lambda_u \sum_i || u_i - h_{+, 0}^{u_i} ||^2 + \lambda_v \sum_j || v_j - h_{+, 0}^{v_j} ||^2 \\
      &+ \lambda_w (\sum_c\sum_l \left(||W_{c,l}^{(u)}||^2 + ||b_{c,l}^{(u)}||^2 \right) \\
      &\qquad + \sum_c\sum_l \left(||W_{c,l}^{(v)}||^2 + ||b_{c,l}^{(v)}||^2 \right) ).
\end{split}
\end{flalign}
\endgroup

The loss function includes reconstruction costs of user and item information sets, the error to predict $r_{i,j}$, and the approximation error between latent factor vectors of feature learning and collaborative filtering. The loss function is minimized to obtain parameters for the DHAs and the CF model. The mean squared error and the negative log-likelihood are used as cost functions for non-sequential and sequential data, separately. We use $\lambda_{m}$, $\lambda_{n}$, $\lambda_u$ and $\lambda_v$ to balance losses between users and items,  $\lambda_f$, and $\lambda_w$ to regularize the weight matrix and bias vectors.


\subsection{Parameter learning}

We apply coordinate descent to alternate the optimization between representation learning of heterogeneous data and user-item interaction, similar to \cite{hwang2015,cwang2011}. Given $W$s and $b$s, the gradients of the loss function $L$ with respect to $u_i$ and $v_j$ are computed and set to 0, leading to the following updates:
\begin{flalign}
u_i &\leftarrow (V^TC_i V +\lambda_f I + \lambda_u I)^{-1}(V^TC_i R_i + \lambda_u h_{+,0}^{u_i}), \\
v_j &\leftarrow (U^TC_j U +\lambda_f I + \lambda_v I)^{-1}(U^TC_j R_j + \lambda_v h_{+,0}^{v_j}),
\end{flalign}
where $U \in \mathbb{R}^{m \times d}$ and $V \in \mathbb{R}^{n \times d}$ contain the user and item latent factor vectors, and $d$ is the vector dimensionality.
Given $U$ and $V$, the weight matrix and bias vectors of every layer are learned by backpropagation with stochastic gradient descent~(SGD). Gradients of $W$ and $b$ are calculated as follows:
\begingroup\makeatletter\def\f@size{8}\check@mathfonts
\begin{flalign}
\frac{\partial L}{\partial W^u} =& \lambda_w W^u + \lambda_m \frac{\partial \bar{S}_c}{\partial W^u} \sum_c \emph{loss}(S_c^u, \bar{S}_c^u) + \lambda_u \frac{\partial h_{+,0}^u}{\partial W^u} (U - h_{+,0}^u), \\
\frac{\partial L}{\partial b^u} =& \lambda_w b^u + \lambda_m \frac{\partial \bar{S}_c}{\partial b^u} \sum_c \emph{loss}(S_c^u, \bar{S}_c^u) + \lambda_u \frac{\partial h_{+,0}^u}{\partial b^u} (U - h_{+,0}^u).
\end{flalign}
\endgroup
A learning rate $\alpha$ is adopted to update all parameters using calculated gradients. 

\section{Experiments}

Experiments are conducted on three real world datasets,  MovieLens-100k~({\em ml-100k}), MovieLens-10M~({\em ml-10m}), and one dataset from an e-commerce company~(OfflinePay). We first investigate whether the flexible autoencoder architecture of our model can generate more accurate latent representations on non-sequential data. Experiments on OfflinePay  evaluate the effectiveness of sequential data modeling.

\subsection{Datasets and preprocessing}

The first dataset, {\em ml-100k}, contains ratings from 943 users on 1,682 movies.
It has demographic data for users and descriptions for movies.
The second dataset, {\em ml-10m}, contains 10,000,054 ratings and 95,580 tags from 71,567 users for 10,681 movies. It  contains item content information, but no demographic data. We employ user-added tags as an information source for users as well as for movies. 
%

OfflinePay is a dataset of user purchases in (offline) shops, paying with a plastic e-money card. The dataset contains a total of 67M transaction records from a four-month period. The goal of using the OfflinePay dataset is to recommend new shop genres to users, not individual products.
After aggregating all transaction data into the format of (user $i$, shop genre $j$, number of transactions $r_{ij}$), and removing shoppers who used only one shop genre, the number of $r_{ij}$ values is 7,150,833 with 961,992 unique users and 105 shop genres. The auxiliary data sources include user registered information and shop genre textual descriptions. Additionally, we collect user purchase history on an e-commerce platform during the same time period. The sequence data contains the genres of purchased items online.

The datasets are preprocessed to fixed-length embeddings for non-sequential data, and sequences of embedding vectors for sequential data, respectively. 
For {\em ml-100k}, we discretize continuous features like age to discrete values, compute a bag-of-words vector for each user and item. The vector dimensions are 821 for users, 2,482 for movies, respectively.
For {\em ml-10m}, movie content description and tags that users give to items are textual information. We first tokenize texts, then train Doc2vec vectors for every data source with the embedding vector length set to 500.

To generate shop genre embedding vectors for the Off\-linePay dataset, all shop names that belong to same genre are grouped together and Doc2vec is applied to generate a 300-dimensional vector for each shop genre. User registered information is preprocessed the same way as {\em ml-100k}, and the vector length is 189. For the sequence of genre purchase history, Word2vec is adopted to build 100-d embedding vectors after tokenization. Genres in each sequence are mapped to the corresponding embedding vectors. 

In experiments, we rank predicted ratings of candidate items and recommend the top $M$ to each user. Mean average precision~(MAP) and recall are used as evaluation metrics.



\begin{figure*}
  \centering
  \hspace*{-2em}
  \begin{subfigure}[b]{0.33\linewidth}\includegraphics[width=\linewidth]{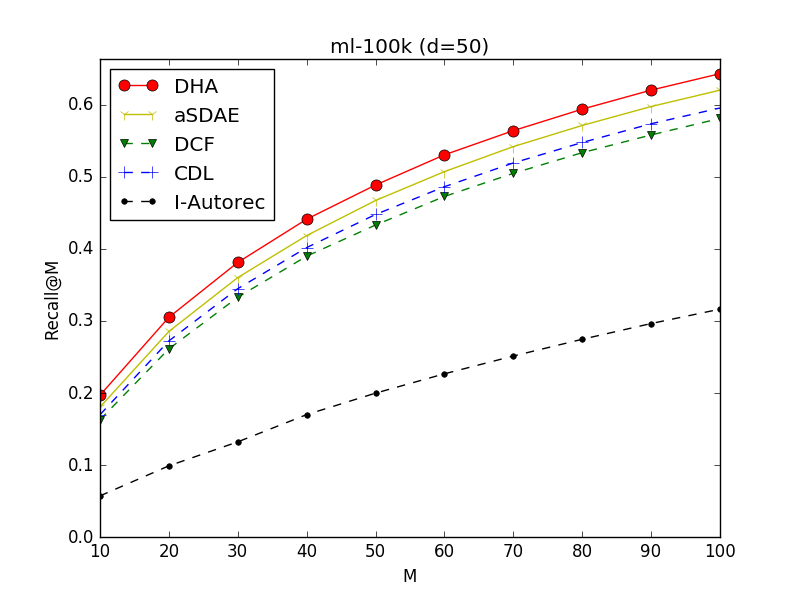} 
  \end{subfigure}\hspace*{-1em}
  \begin{subfigure}[b]{0.33\linewidth}\includegraphics[width=\linewidth]{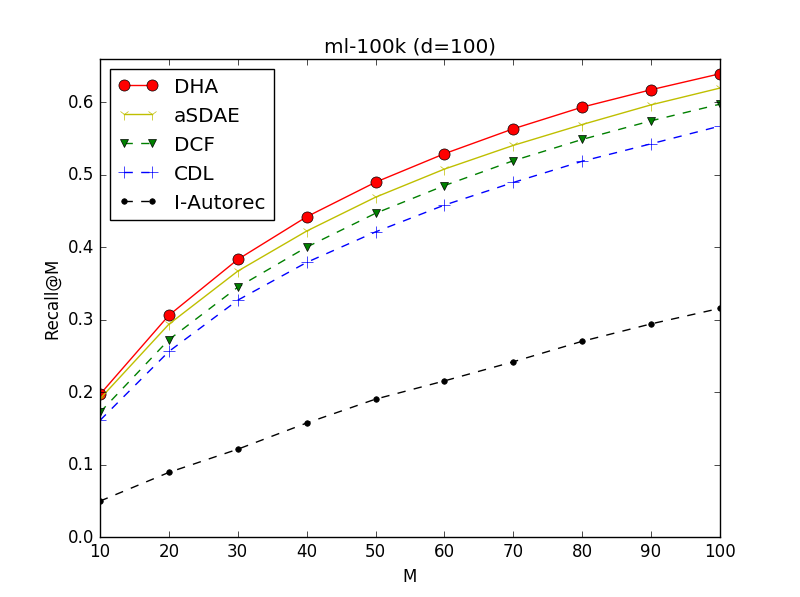}
  \end{subfigure}\hspace*{-1em}
  \begin{subfigure}[b]{0.33\linewidth}\includegraphics[width=\linewidth]{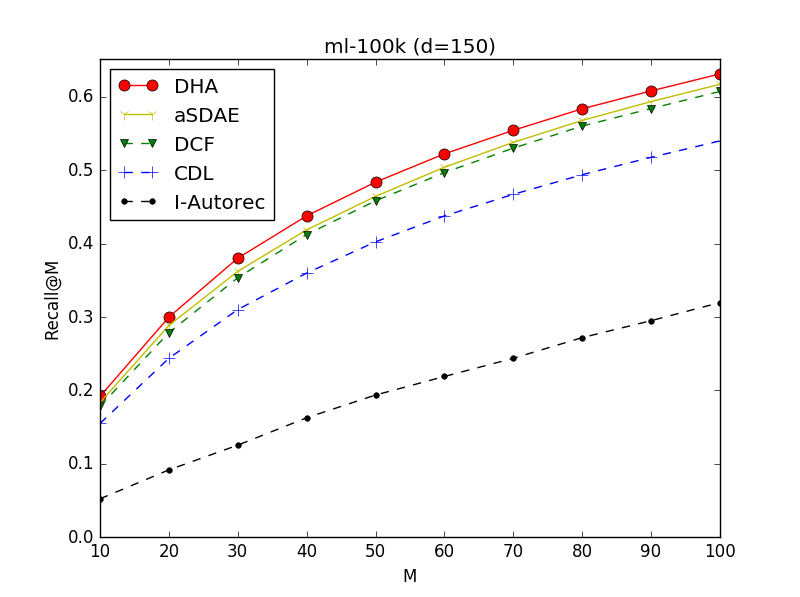}
  \end{subfigure}\hspace*{-1em}
  
  \hspace*{-2em}
  \begin{subfigure}[b]{0.33\linewidth}\includegraphics[width=\linewidth]{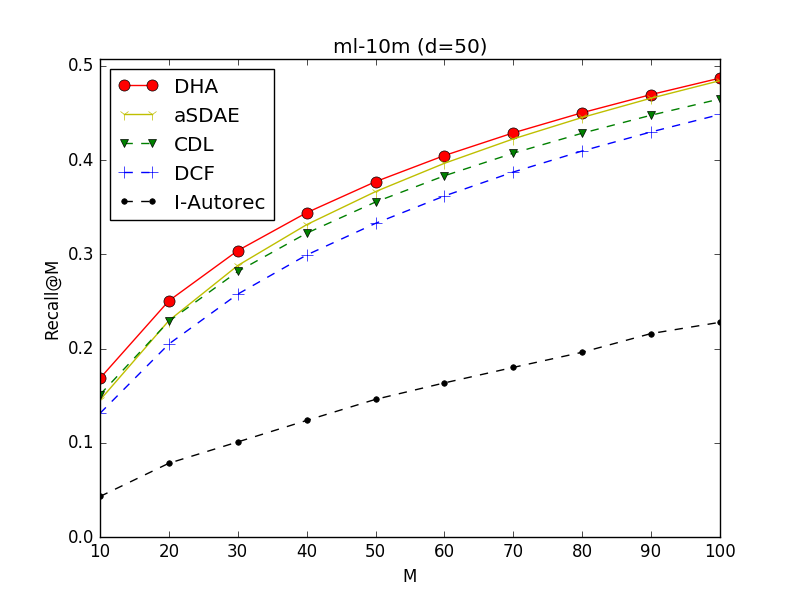}
  \end{subfigure}\hspace*{-1em}
  \begin{subfigure}[b]{0.33\linewidth}\includegraphics[width=\linewidth]{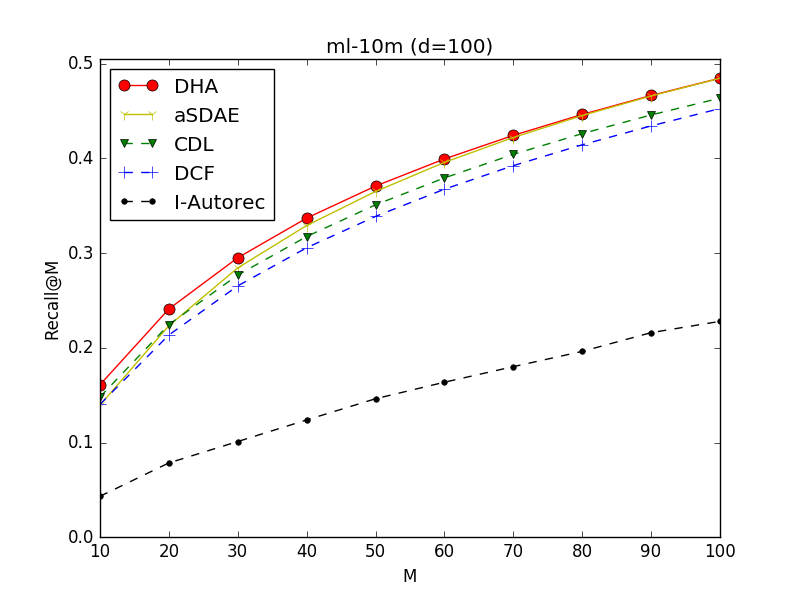}
  \end{subfigure}\hspace*{-1em}
  \begin{subfigure}[b]{0.33\linewidth}\includegraphics[width=\linewidth]{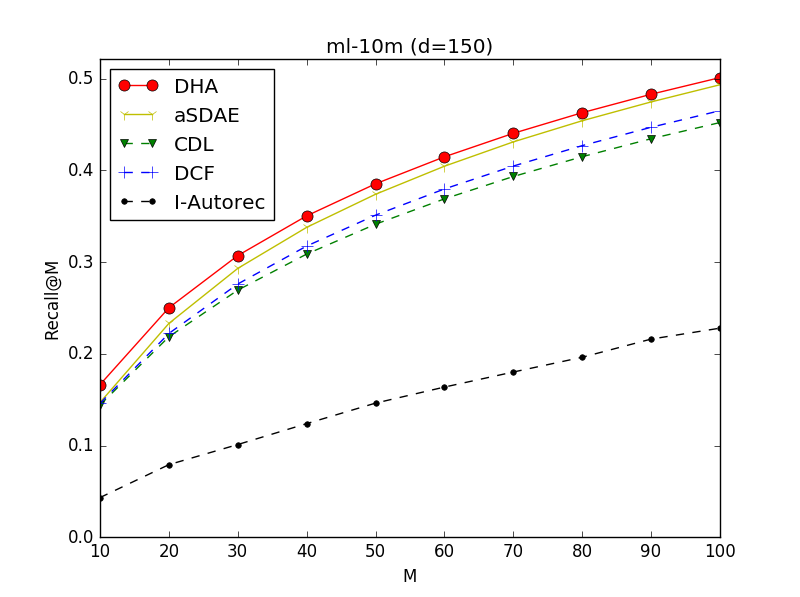}
  \end{subfigure}\hspace*{-1em}
  
  \caption{{\bf Recall@M comparison on {\em ml-100k, ml-10m}.} \it Results are shown for I-AutoRec, CDL, DCF, aSDAE and DHA. The dimension~$d$ of the latent factor vector is set to 50, 100 and 150, respectively.}
  \label{fig:recall_comparison_ml-100k_and_ml-10m}
\end{figure*}

\subsection{Experimental setting}




The number of hidden layers of each model is optimized on a validation dataset.
The first fusion hidden layer of DHA is used to bridge the joint training between feature space learning and collaborative filtering. For other models, if the total number of hidden layers is $L$, we connect layer $L/2$ for joint training. The number of units in each hidden layer is incremented by $K$ from the middle of the autoencoder to both sides.
For sequential data modeling, recent $T$ purchases is used in the experiments, and values $T \in \{5, 10\}$ are evaluated in our experiments.

The mini-batch size is set to 50 and 1,000 for {\em ml-100k} and {\em ml-10m}, respectively. For the OfflinePay dataset, since the numbers of unique users and items differ significantly, it is set to 20 for items and 10,000 for users, separately.
The model is implemented using the Theano library. 

\subsection{Experiments on MovieLens datasets}
\newcommand{\bftab}{\fontseries{b}\selectfont}

We compare our model with the following algorithms. Note that experiments on {\em MovieLens} do not include sequential data.


\begin{itemize}
\item AutoRec\cite{Sedhain2015}: I-AutoRec takes a partial item feedback vector as input and reconstructs at the output layer.

\item CDL\cite{hwang2015}: a hierarchical Bayesian model that jointly performs deep representation learning for content information and collaborative filtering for the ratings matrix.

\item DCF\cite{Lis2015}: a model that combines matrix factorization with marginalized denoising stacked autoencoders. We concatenate side information as input to DCF.

\item aSDAE\cite{Dong2017}: a hybrid model that integrates side information by an additional denoising autoencoder into the matrix factorization model.
\item DHA: the proposed model that applies independent autoencoder architecture to heterogeneous data sources.
\end{itemize}

To compare different models, we repeat 80-20 splits of the data 5 times, run 5-fold cross validation and report average performance.
Grid search is applied to find optimal hyperparameters for all models. We search the learning rate of SGD, $\alpha \in \{ 0.1, 0.05, 0.01, 0.001 \} $, the regularization of learned parameters, $\lambda_f$ and $\lambda_w$ of our model $ \in \{ 2.0, 0.1, 0.01, 0.001 \} $, the corruption level of masking noise $\in \{ 0.1, 0.3 \} $, the activation function $\in \{sigmoid, relu\} $, and the number of fusion hidden layers $\in \{ 1, 2 \} $. The parameters used to balance loss between user and item, $\lambda_{m}, \lambda_{n}, \lambda_{u}, \lambda_{v}$ are set to 1. For CDL, DCF and aSDAE, we search hidden layer number from 4 and 6. 
The joint training is alternated 5 times, and we run 5 epochs for learning features in each alteration. Before the joint training, layer-wise pretraining is conducted to initialize network weights.

For the experiment on {\em ml-100k}, we input rating vectors, item content information and user demographic data to DHA and aSDAE.
Rating vectors are not used in DCF and only item content information is used in CDL. I-AutoRec leverages no side information. After grid search, the adopted hidden layer number of CDL, DCF and aSDAE is 4. The number of fusion hidden layer is set to 1 for DHA. The parameter for regularizing learned parameters is set to 0.01 in DHA, 0.001 in CDL and aSDAE, and 0.1 in DCF, respectively. The optimal performance is found when the learning rate is set to 0.001 for CDL, DCF, 0.01 for aSDAE and DHA, and 0.1 for I-Autorec.

\begingroup
\setlength{\tabcolsep}{5pt} 
\begin{table}
\caption{{\bf MAP@100 comparison on {\em ml-100k} and {\em ml-10m} datasets.} \it Results are shown for three different settings of user and item latent factor vectors, $d$=50, 100, and 150.}
\begin{center}
\label{tab: map@100 for ml-100k, ml-10m}
\begin{tabular}{ l c c c c  c c c c}
\toprule
 & \multicolumn{4}{c}{{\em ml-100k}}  & \multicolumn{4}{c}{{\em ml-10m}} \\
\cmidrule{3-5} \cmidrule{7-9}
Model && d=50 & d=100 & d=150 && d=50 & d=100 & d=150 \\
\cmidrule{1-2} \cmidrule{3-5} \cmidrule{7-9}
I-AutoRec && 0.0573 &  0.0568 & 0.0572 && 0.0325  &  0.0323 & 0.0326 \\ 
CDL && 0.1896 & 0.1825 & 0.1685 && 0.1458 & 0.1532 & 0.1612 \\
DCF && 0.2012 & 0.2028 &  0.2069 && 0.1591 & 0.1620 &  0.1566 \\
aSDAE && 0.2161 & 0.2228 &  0.2142 && 0.1602  & 0.1560  &  0.1642 \\ 
 \bftab DHA && \bftab 0.2236 & \bftab 0.2304 & \bftab 0.2258 &&\bftab 0.1793 & \bftab 0.1774  & \bftab 0.1824\\ 
\bottomrule
\end{tabular}
\end{center}
\end{table}
\endgroup

As shown in Fig. \ref{fig:recall_comparison_ml-100k_and_ml-10m}, all models achieve better recall than I-AutoRec, showing the advantage of using side information. DHA and aSDAE perform better than CDL which only incorporates item content description. DHA outperforms aSDAE which integrates raw side information at every hidden layer. 
The MAP comparison in Table \ref{tab: map@100 for ml-100k, ml-10m} shows our model obtains more precise results for all dimension settings. 

There are five sets of available inputs for the experiment on the {\em ml-10m} dataset. Users and movies have rating and tag vectors, movies also have content vectors. 
For CDL, DCF and aSDAE, different information vectors are concatenated  as input. Our model uses all components, {\it i.e.} two components for users and three components for movies. In the experiment, the best performance is obtained when the number of hidden layers is set to 4 for CDL, aSDAE and to 6 for DCF. In our model, we use 2 fusion hidden layers and different layer numbers for components. The number of hidden layers, $L_c$, is set to 4 for users and movie rating vectors and to 2 for tag and content vectors. 
As shown in Fig. \ref{fig:recall_comparison_ml-100k_and_ml-10m}, DHA obtains better recall performance compared to other algorithms. aSDAE is competitive and outperforms both DCF and CDL in three dimension settings. 
The MAP comparison in Table \ref{tab: map@100 for ml-100k, ml-10m} indicates that in addition to producing recommendation with better recall, our model also achieves better precision results.

\begin{figure*}
  \centering
  \hspace*{-2em}
  \begin{subfigure}[b]{0.34\linewidth}
  	\includegraphics[width=\linewidth]{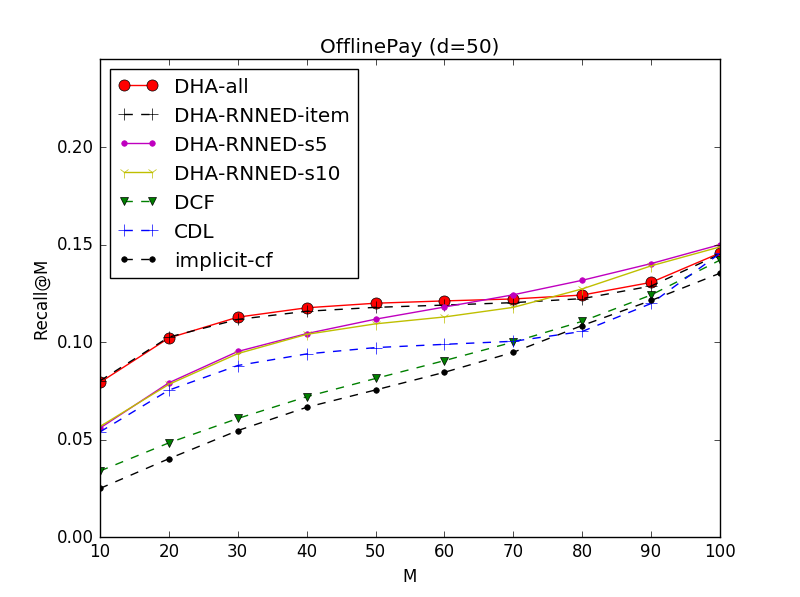}
  \end{subfigure}\hspace*{-1em}
  \begin{subfigure}[b]{0.34\linewidth}
  	\includegraphics[width=\linewidth]{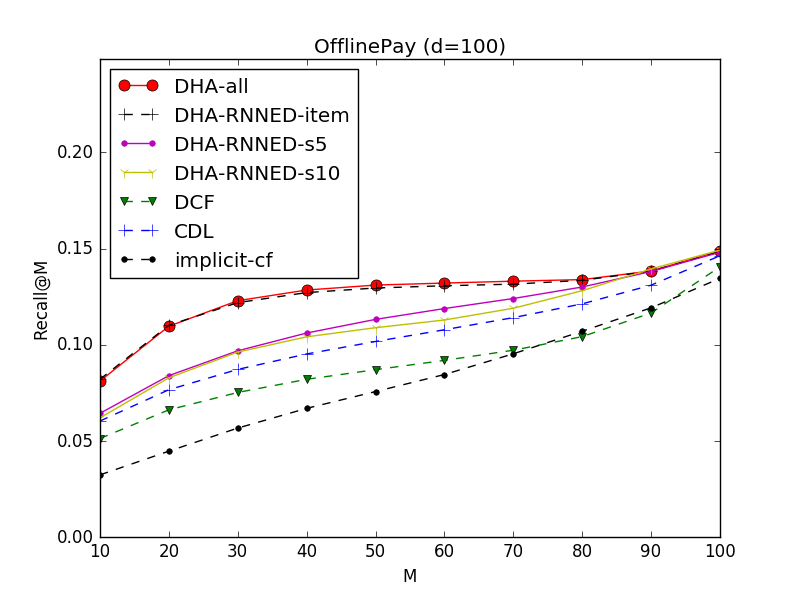}
  \end{subfigure}\hspace*{-1em}
  \begin{subfigure}[b]{0.34\linewidth}
  	\includegraphics[width=\linewidth]{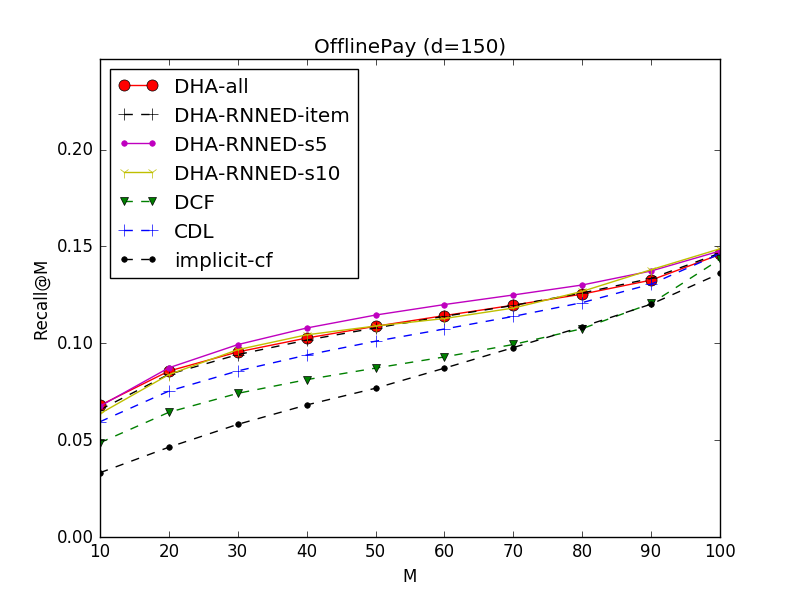}
  \end{subfigure}\hspace*{-1em}
  \caption{{\bf Recall@M comparison on {\em OfflinePay} dataset.} \it Results are shown for implicit-CF, CDL, DCF, DHA-RNNED-s10, DHA-RNNED-s5, DHA-RNNED-item and DHA-all. The dimension of the latent factor vector is set to 50, 100 and 150, respectively.}
  \label{fig:recall_comparison_offlinepay}
\end{figure*}

\subsection{Experiments on OfflinePay dataset}

Since the OfflinePay dataset involves user online purchase histories, we use the first 3-month data as training data, the following half month's data as validation dataset to find optimal parameters, and data from the remaining half-month as test set. 
We compare the following algorithms:

\begin{itemize}
\item implicit-cf\cite{Hu2008}: a matrix factorization model for implicit datasets.

\item CDL\cite{hwang2015}: a Bayesian model that learns a feature space from item information and jointly trains with CF.

\item DCF\cite{Lis2015}: a model that incorporates side information by marginalized denoising stacked autoencoders with a matrix factorization model. 

\item DHA-RNNED-s10: our model that learns a latent representation only from the sequence of online purchases. The number of time steps in each sequence is 10.

\item DHA-RNNED-s5: our model with the same modeling process as DHA-RNNED-s10, but using 5 time steps in each sequence.

\item DHA-RNNED-item: our model extracts features from sequential online purchases at user side, and from shop genre descriptions at item side.

\item DHA-all: the proposed model that leverage non-sequential side information sets and sequential online purchase activities simultaneously. The used time step number of the purchase sequence is 10.
\end{itemize}

In the experiment, the joint learning is alternated 3 times, and we run 3 epochs for feature extraction every time. The number of hidden layers for CDL, DCF and our model is set to 4, and 1 fusion hidden layer is used in DHA models. 
For the sequential modeling, we set the hidden units of LSTMs to be the same as the dimension of the user and item latent factor vector. The SGD learning rate and regularization parameters for each model are found by grid search on the validation set. We set the learning rate to 0.1 for implicit-cf and CDL, to 0.001 for DCF and to 0.01 for the other models. The parameter to regularize learned parameters is set to 2.0 for CDL, and 0.1 for DCF and DHA-all. 
There is no training alteration for implicit-cf, but we run 25 iterations to learn user and item latent factor vectors.

\begin{table}[b]
\caption{{\bf  MAP@100 comparison on {\em OfflinePay} dataset.} \it Results are shown for different dimensions of the latent factor vector, $d$=50, 100, 150.}
\label{tab: map@100 for offlinepay}
\begin{center}
\resizebox{.8\columnwidth}{!}{
\begin{tabular}{ l c c c }
\toprule
Models & d=50 & d=100 & d=150 \\
\midrule
implicit-cf & 0.0155 &  0.0178 & 0.0177  \\ 
CDL & 0.0296 & 0.0336 & 0.0333  \\
DCF & 0.0237 & 0.0311 &  0.0306   \\ 
\midrule
DHA-RNNED-s10 & 0.0327 & 0.0333 &  0.0339  \\
DHA-RNNED-s5 & 0.0307 & 0.0343 & \bftab 0.0367  \\ 
DHA-RNNED-item & 0.0394 & 0.0402 & 0.0345  \\
DHA-all & \bftab 0.0424 & \bftab 0.0403 & 0.0361 \\ 
\bottomrule
\end{tabular}}
\end{center}
\end{table}


DCF integrates both user registration information and shop genre descriptions, while CDL uses only the latter one. DHA-RNNED-s10 and DHA-RNNED-s5 do not include any side information except user online purchases. DHA-RNNED-item adopts sequential data and shop genre descriptions, and DHA-all utilizes all of the data. Note that since ratings are not used in any models, aSDAE is not applied on OfflinePay dataset. 

From Fig.~\ref{fig:recall_comparison_offlinepay}, we observe that models taking advantage of side information have better recall than the baseline implicit-cf. 
CDL outperforms DCF which, in fact, uses more information sets. This may be due to the fact that many user registration records have outdated or missing values, making the feature extraction less accurate.
Compared to CDL and DCF, the proposed models with sequential data modeling achieve better recall. This is due to the fact that offline shop genres in the dataset are included in the online purchased genres. This also indicates that the latent features is able to be extracted from recent online purchases accurately, and reflect the trends of user interests, then lead to better recommendations for offline products, as well. 
The MAP comparison in Table \ref{tab: map@100 for offlinepay} shows that the models involving sequential modeling achieve higher precision. This consistently shows that the modeling of online purchases helps with offline product recommendation.


The recall comparison in Fig.~\ref{fig:recall_comparison_offlinepay} shows that DHA-RNNED-s10 and DHA-RNNED-s5 have a similar trend as recommended item $M$ increases. These two models use only the sequence of purchased genres from an online e-commerce platform, but with different time steps in the sequence. it is also shown that DHA-all and DHA-RNNED-item have similar recalls. The difference between these two models is that the latter model does not include user registered data. Linking to the previous observation that CDL outperforms DCF, user data does not significantly contribute to the recommendation results.

In order to compare the effect of purchase recency of the input sequence, we apply DHA-RNNED-s10 and DHA-RNNED-s5 to encode the recent ten and five purchases, respectively. Our hypothesis is that more recent online purchases are more representative of current user interests. Although the difference is not big, the recall and MAP comparisons support our hypothesis. 
The experiments demonstrate that with the independent autoencoder structure for user and item side information and the modeling of user online activities, our model is able to achieve competitive recall and MAP results.


\section{Conclusions}

We proposed a model that incorporates multiple sources of heterogeneous auxiliary information in a consistent way to alleviate the data sparsity problem of recommender systems. It takes static and sequential data as input and captures both the inherent tastes of users as well as the dynamics of user preference. The model uses a flexible autoencoder structure for integrating different data sources leading to significant performance gains.



\end{document}